\documentclass[conference, 9pt]{IEEEtran}
\IEEEoverridecommandlockouts

\usepackage{hhline}
\usepackage{colortbl}
\usepackage{centernot}
\usepackage{pbox}
\usepackage{amsmath}
\usepackage{amsfonts}
\usepackage{url}
\usepackage{bm}
\usepackage{comment}
\usepackage{graphicx}
\usepackage{setspace}
\usepackage{epstopdf}
\usepackage{multirow}
\usepackage{color}
\usepackage{tabu}
\usepackage{mdframed}
\usepackage{syntax}
\usepackage{fancyvrb}
\usepackage{subfig}
\usepackage[hidelinks]{hyperref}
\usepackage{cleveref}
\crefname{figure}{fig.}{figures}
\Crefname{figure}{Fig.}{Figures}
\Crefname{table}{TABLE}{Tables}

\usepackage{paralist}
\usepackage{stmaryrd}
\usepackage{tikz}
\usepackage[referable]{threeparttablex}
\usepackage{tabularx}
\usepackage{booktabs}
\usepackage{array}
\usepackage{fancyhdr}
\usepackage{float}
\usepackage{xspace}
\usepackage{pifont}
\usepackage{balance}
\usepackage{xcolor}
\usepackage{tablefootnote}
\usepackage[normalem]{ulem}

\usepackage{tikz}
\usepackage{xcolor}
\newcommand*\circled[1]{\tikz[baseline=(char.base)]{
            \node[shape=circle,fill,inner sep=1pt,scale=0.8] (char) {\textcolor{white}{#1}};}}

\usepackage{graphicx}
\usepackage{adjustbox}

\newfloat{figtab}{htb}{fgtb}
\makeatletter
  \newcommand\figcaption{\def\@captype{figure}\caption}
  \newcommand\tabcaption{\def\@captype{table}\caption}
\makeatother

\usepackage{algorithm} 
\usepackage[noend]{algpseudocode}

\definecolor{princetonorange}{RGB}{255,143,0}

\definecolor{lightgreen}{RGB}{198, 224, 183}

\definecolor{lightred}{RGB}{240, 205, 176}

\usepackage{soul}
\newcommand{\yao}[1]{\textcolor{black}{#1}}

\usepackage[left=1.5cm, right=1.5cm, top=1.376cm, bottom=1.8cm]{geometry}

\begin{document}
\setstretch{0.925}





\vspace{-.3in}
\title{SynCircuit: Automated Generation of New \underline{Syn}thetic RTL \underline{Circuit}s Can Enable Big Data in Circuits}




\author{Shang Liu$^\dagger$, Jing Wang$^\dagger$, Wenji Fang, Zhiyao Xie$^\ast$\\
Hong Kong University of Science and Technology\\
\{sliudx, jwangjw, wfang838\}@connect.ust.hk, eezhiyao@ust.hk

}

\vspace{-.3in}
\maketitle
\vspace{-.2pt}
{\renewcommand{\thefootnote}{}
 \footnotetext{$\dagger$ Equal Contribution}
 \footnotetext{$\ast$ Corresponding Author}}
\vspace{-.3in}
\vspace{-.2pt}

\vspace{-.5in}
\begin{abstract}

In recent years, AI-assisted IC design methods have demonstrated great potential, but the availability of circuit design data is extremely limited, especially in the public domain. The lack of circuit data has become the primary bottleneck in developing AI-assisted IC design methods. In this work, we make the first attempt, SynCircuit, to generate new synthetic circuits with valid functionalities in the HDL format.

SynCircuit automatically generates synthetic data using a framework with three innovative steps: 1) We propose a customized diffusion-based generative model to resolve the Directed Cyclic Graph (DCG) generation task, which has not been well explored in the AI community. 2) To ensure our circuit is valid, we enforce the circuit constraints by refining the initial graph generation outputs. 
3) The Monte Carlo tree search (MCTS) method further optimizes the logic redundancy in the generated graph. Experimental results demonstrate that our proposed SynCircuit can generate more realistic synthetic circuits and enhance ML model performance in downstream circuit design tasks.

\end{abstract}




\section{Introduction}
\label{sec:intro}



The ever-increasing demands for chip performance have caused escalating integrated circuit (IC) complexity, challenging traditional Electronic Design Automation (EDA) methodologies. 
In recent years, AI-assisted IC design techniques have demonstrated remarkable potential in accelerating the chip design process. Notable AI for EDA applications include automated chip design generation~\cite{peibetterv, chang2024lamagic, liu2023rtlcoder}, fast chip quality prediction~\cite{bai2023towards, fang2023masterrtl, xu2022sns, qin2024cross, yang2022versatile, chai2023circuitnet, zou2024circuit}, and automated chip design planning~\cite{bai2023archexplorer}.


\textbf{SynCircuit: Pathway to Big Data in Circuits.} Compared with generating datasets with limited circuits in the public domain, we believe the automated generation of a large number of \emph{synthetic} circuits is the most promising way to \emph{completely solve} the circuit data availability problem in the foreseeable future. In this work, we demonstrate the feasibility of this promising direction with a novel synthetic circuit generation framework named SynCircuit. 


To the best of our knowledge, SynCircuit proposed in this work is the first technique to automatically generate brand-new synthetic circuits with valid functionalities. Specifically, we achieve this with novel customized graph learning algorithms and follow-up refinement techniques. 
The synthetic circuits are in the RTL stage and support hardware description language (HDL) code format. They can be automatically synthesized into regular netlists and turned to layouts with commercial tools. 
These synthetic designs can not only enhance the training and robustness of AI models by providing diverse and extensive datasets but also may serve as benchmarks for assessing the performance of existing design algorithms~\cite{yoon2022graph, chai2023circuitnet, fang2025cfm}. In addition, our proposed directed graph generation framework potentially applies to other tasks, such as neural architecture search (NAS) and Bayesian optimization~\cite{jiang2020efficient,zhang2019d, dudziak2020brp}.

SynCircuit generates synthetic circuits with 3 phases: 

\textbf{Phase 1. Generation of Large Directed Cyclic Graph (DCG).} Circuit design code (i.e., HDL code) can be equivalently represented as directed cyclic graphs (DCG), 
allowing the utilization of graph generation algorithms to create synthetic circuit designs.
However, most machine learning (ML) research on graph generation has focused on undirected graphs~\cite{you2018graphrnn, jiang2020efficient, simonovsky2018graphvae, liao2019efficient, chen2023efficient}, leaving unsolved challenges on generating \emph{directed graphs}. 
For instance, edge predictions in these prior works are determined by applying a \emph{symmetric operator} to the pairwise node embeddings.
Also, spectrum-based generation techniques like Spectre~\cite{martinkus2022spectre} and DiGress~\cite{vignac2022digress} rely on properties specific to undirected graphs, such as \emph{symmetric Laplacian matrices}. 
Some works generate directed graphs in an autoregressive manner following the topological order~\cite{zhang2019d, li2024layerdag}. However, they do not apply to directed \emph{cyclic} graphs, where a topological sort does not exist. In summary, the generation of DCG is not a well-solved problem, not only for synthetic circuit data generation in EDA applications, but also in the general AI community.

In response to the challenges posed by large directed cyclic graphs in circuit designs, we propose a graph generation algorithm for digital circuits. We employed a diffusion generation framework for graph generation and designed an efficient and low-cost denoising network that includes an encoder and a decoder. For the graph encoder, a directed messaging-passing neural network is utilized to better capture the local features of graphs while only requiring a relatively low memory consumption. For the decoder, we model the directed edge relation based on learnable translated embeddings.

\begin{figure}[!t]  
    \centering        
    \includegraphics[width=0.49\textwidth]{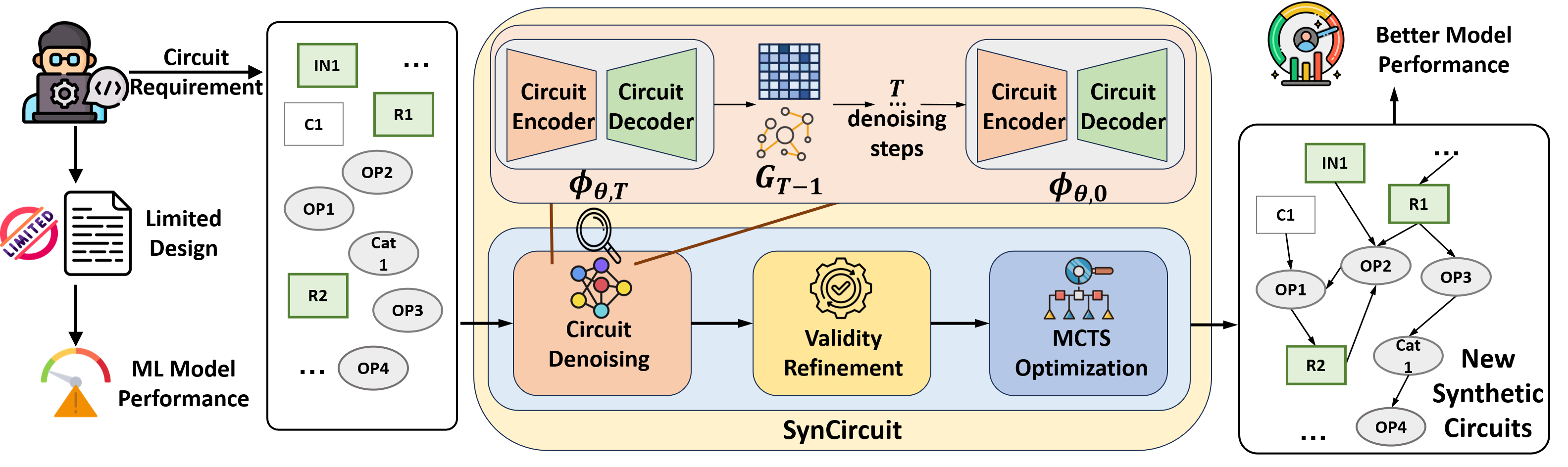}  
    \caption{The overview of SynCircuit.
     It addresses the serious design scarcity issue by generating new synthetic circuits. Through a three-stage process consisting of denoising, validity refinement, and optimization, SynCircuit can generate an unlimited number of synthetic circuits suitable for downstream ML tasks.} 
    \vspace{-0.1in}
    \label{fig:intro}  
\end{figure}

\textbf{Phase 2 \& 3. Refinement of Synthetic Circuits.} Additionally, how to ensure the graph validness, i.e., satisfying the circuit constraints, is not also fully explored in graph generation,  especially in a one-shot manner. Given the fanin constraints and combinational loop prevention requirements, we propose a probability-guided graph post-processing strategy to further refine the invalid synthetic graph obtained from the denoising process. This technique sequentially processes the fan-in edges of each node, which not only ensures that the resulting circuit satisfies the constraints but also preserves the generative information of the diffusion model as much as possible.

We also point out the logic redundancy problem in the synthetic circuits. To further alleviate the redundancy of the generated designs and improve their utility as data augmentation in downstream tasks, we also employ Monte Carlo Tree Search (MCTS) to further fine-tune the generated graphs. This approach allows us to obtain higher-quality graphs that are better suited for real applications.


We summarize our contributions as follows:
\begin{itemize}
    \item To the best of our knowledge, we are the first to propose a framework for generating digital circuits at the RTL level from a graph generation perspective. This framework comprises three components in the generation phase: directed cyclic graph generation, probability-guided postprocessing, and MCTS-based optimization.
    \item We propose a graph generative model solution targeting the challenging directed cyclic graph generation task. We utilize the diffusion framework and reconstruct the edge direction information through an asymmetric edge decoder.
    \item To further alleviate the redundancy of the generated designs and improve their utility as data augmentation in downstream tasks, we employ MCTS to further fine-tune the generated graphs. This approach obtains higher-quality graphs with optimized logic redundancy which are better suited for real applications.
    \item We conducted experiments on graph structural property similarity evaluation and demonstrated that SynCircuit generates more realistic graphs. Experiments on early RTL-stage PPA prediction show that SynCircuit can help alleviate the data availability problem in AI-based solutions for EDA tasks. 
\end{itemize}


\section{Problem Formulation}

\textbf{DCG representation of HDL code.} Given a circuit design HDL code $D$, it can be mapped to a circuit graph representation $G$ through a bijection function $f: D \leftrightarrow  G$ using our developed parser. Here, $G$ is a directed cyclic graph represented as $(V, E, X)$, where $V$ denotes the set of nodes, $E$ represents the set of edges with $e_{i,j} \in E$ indicating a directed edge from $v_i$ to $v_j$, and $X$ represents the attributes of the nodes. The node attributes include node \textit{type} and \textit{width}. The node types are primarily categorized into IO port, arithmetic operator, register (reg), bit selection, and concatenate operator. The width attribute reflects the output signal width of the corresponding node. 


\textbf{Circuit constraints $\mathcal{C}$.} To ensure that the graph can be parsed back into HDL code, valid $G$ must satisfy two types of constraints: 
\begin{itemize}
    \item The node type uniquely determines the number of parent nodes. For example, a node of the type ``mux" requires three parent nodes, while the type ``add" requires two.
    \item The graph must not contain any combinational loops. The combinational loop is defined as a cycle that does not include any registers, which would cause timing violations.
\end{itemize}


\textbf{Generate new large circuits.} In order to provide generation flexibility, we will use the generative model to produce edges $E$ conditioned on the specified node number $V$ and attributes $X$ by users. Our objective is to learn the probability distribution $P(G\mid V, X)$ from a set of real circuit HDL codes through their graph representations $\left \{  G_i\right \} $. This will enable the generation of a series of new synthetic circuit graphs $\left \{  G^{syn}_i\right \} $.  We aim for these synthetic circuit design graphs to closely resemble real designs in terms of topological structure and satisfy all the circuit constraints $\mathcal{C}$. Moreover, considering most registers in real designs will not be removed in the synthesis stage, we require the generated circuits to have as low logic redundancy as possible. 
\section{Overview of SynCircuit}

\begin{figure*}[t]  
    \centering        
    \includegraphics[width=0.95\textwidth]{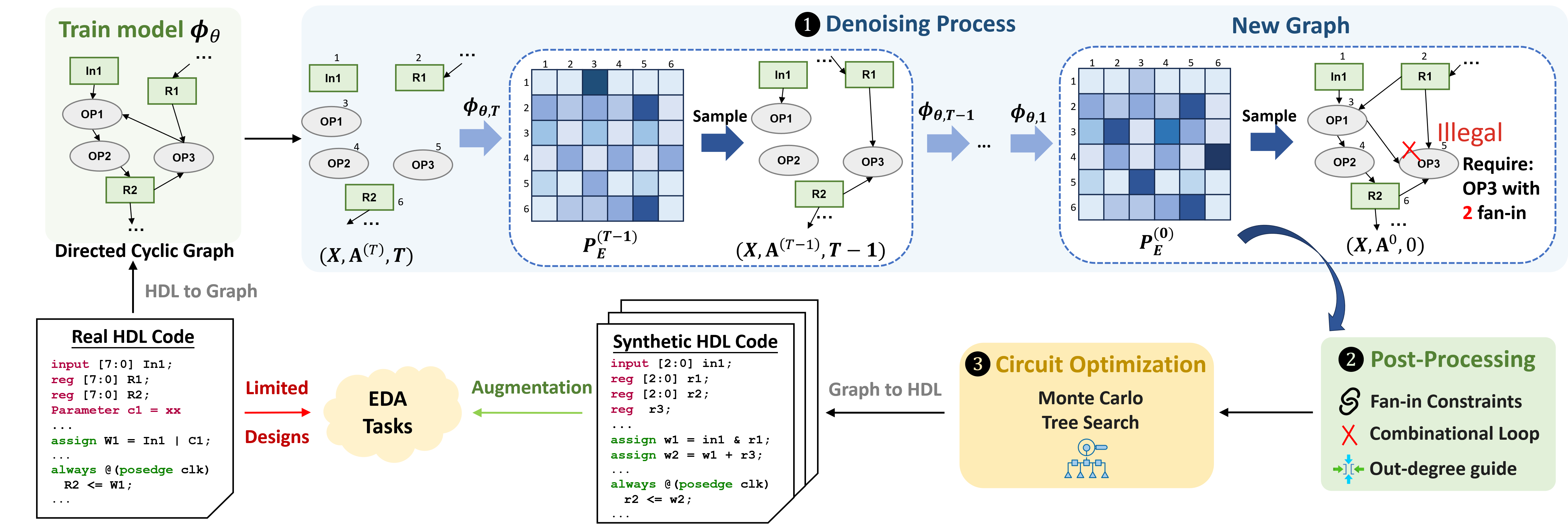}  
    \caption{Overview of SynCircuit framework. The circuit generation process starts with the specified node number $V$ and attributes $X$. In \textbf{Phase 1}, the model performs a reverse diffusion process, where it predicts the probability of an edge existing between any pair $(i, j)$ to obtain $P_E^{(t)}$. Then we sample from $P_E^{(t)}$ to update a graph $G^{(t)}$ as a new time state. After $T$ denoising steps, the $G^{ini}=G^{(0)}$ is obtained. However, the generated graph may not necessarily be valid.  In \textbf{Phase 2}, during post-processing, we sequentially assign parents to each node based on $P_E^{(t=0)}$, ensuring at each step that the circuit satisfies the constraints. In \textbf{Phase 3}, considering that some synthetic circuits exhibit significant logic redundancy, we employ an MCTS-based optimization technique, thereby reducing the generated design redundancy and better serving the downstream tasks.}  
    \vspace{-.1in}
    \label{fig:Framework-flow}  
\end{figure*}

Initially, the realistic designs are converted into directed cyclic graph representation. We subsequently trained a diffusion model $P(G\mid V, X)$ on these designs. In the generation stage, as Figure~\ref{fig:Framework-flow} shows, our proposed circuit generation process consists of three phases: initial graph generation, validness post-process, and optimization refinement. 
 \vspace{-0.1in}
 $$P({G})\overset{ \circled{1}}{\rightarrow}  G^{ini}\overset{\circled{2}}{\rightarrow}G^{val}\overset{\circled{3}}{\rightarrow}G^{opt}$$
  \vspace{-0.2in}
  
 In the generation phase of \circled{1}, we adopt the trained diffusion model and apply the denoising process to obtain an initial synthetic graph $G^{ini}$ with a corresponding edge probability matrix $P_E^{(t=0)}$. The value of $(i, j)$ in $P_E$ represents the directed edge $e_{i,j}$ existence probability.
 In phase \circled{2}, we propose an autoregressive method to determine the edges based on the $G^{ini}$ to ensure that the refined graph $G^{val}$ satisfies all the circuit constraints $\mathcal{C}$. In the final phase \circled{3}, we address the issue of significant logical redundancy in the generated circuits. $G^{val}$ is fine-tuned by the MCTS-based method, reducing logic redundancy and thereby enhancing the performance of downstream tasks. In the following subsections, we will explain each stage in detail.



\section{Phase 1. DCG Generation Technique}
\label{sec:method.gen}

 \textbf{\yao{Limitation of prior works on DCG} generation.} 
 \yao{Many prior graph generation works utilize a symmetric operator or semi-positive-definite Laplacian matrix to assign edges.}
 This symmetric assumption \yao{does not support} the asymmetric edge in the directed cyclic graph. For directed acyclic graphs (DAGs), previous work~\cite{li2024layerdag, zhang2019d} addresses DAG generation by sorting node order based on the topological structure and then generating nodes in a layerwise manner. Topological sorting ensures that a node's parent nodes always precede it in the order. Consequently, the edge direction is automatically determined. However, there is no topological node ordering in the DCG graph since the nodes in a loop share the same topological level.


\textbf{The proposed DCG generation technique overview.} To the best of our knowledge, we are the first to address the task of generating directed cyclic graphs using a diffusion model framework. The method involves a forward diffusion process and a reverse denoising process. In the \textbf{forward process}, we gradually corrupt the adjacency matrix $A$ by adding noise, transforming it from the original data distribution into a noisy distribution. The goal of the reverse process is to reconstruct the original adjacency matrix from the noisy version by reversing this corruption.
During the \textbf{reverse denoising process}, \yao{a} denoising neural network $\phi_{\theta}$ with an \emph{encoder} and a \emph{decoder} is trained with real circuits. The \emph{encoder} processes the node attributes and the noisy adjacency matrix to generate node representations that \yao{consist of rich} graph structural information at each denoising step. The \emph{decoder} utilizes these node representations to effectively reconstruct the ground truth edge connections. By training $\phi_{\theta}$ in the reverse process, the model learns to progressively denoise the graph structure, ultimately generating synthetic circuits similar to realistic ones.

\subsection{{Forward Process}}


 In the forward diffusion process, we progressively corrupt the adjacency matrix $A$ by applying a series of predefined transition matrices, introducing noise at each time step $t$. Specifically, starting from the initial graph $G^{(0)} = (X, A^{(0)})$, the corrupted adjacency matrix $A^{(t)}$ at time $t$ is obtained via a Markov process: 
\[
A^{(t)} = A^{(t-1)} \tilde{Q}_A^{(t)}
\]
where $\tilde{Q}_A^{(t)}$ is the \yao{noise} transition matrix at time $t$. By iteratively applying these transitions, we gradually transform $A^{(0)}$ into a noisy distribution. To control the degree of corruption, we employ a scheduling strategy (e.g., cosine schedule~\cite{nichol2021improved}) to adjust the noise level over time. The scheduling ensures that the corrupted adjacency matrices \yao{$A^{(t)}$} smoothly \yao{transit} from the original data distribution to a predetermined noise distribution.

\subsection{{Denoising Process}}


In the denoising diffusion process, our goal is to recover the original adjacency matrix $A^{(0)}$ from the noisy matrix\footnote{In training, the $A^{(T)}$ is obtained from the forward process. In inference, we randomly sample a completely noisy adjacency matrix} $A^{(T)}$. 
Based on the node attributes\footnote{In training, the $X$ is obtained directly from the real circuits. In inference, we can either use the $P(X)$ distribution from the real design or set it according to the user's specifications.} $X$,  we model the conditional probability $p_\theta(G^{(t-1)}) = p_\theta(A^{(t-1)} | A^{(t)}, X, t)$, parameterized by a denoising network $\phi_{\theta,t}$. In the generation stage, the $\phi_{\theta,t}$ will reconstruct a probability matrix $P_E^{t-1}$ given $G^{(t)}$. The new graph $G^{t-1}$ will be updated by sampling the edge existence probability matrix $P_E^{t-1}$.


\subsection{{Encoder Design for Large Graphs}}
In the denoising network architecture $\phi_{\theta,t}$, the encoder is to capture the structural information of the input noisy \yao{graph} $G^{(t)}$.
Handling large-scale graphs with more than 10K nodes poses computational challenges. To address this, we design an efficient encoder based on Message Passing Neural Networks (MPNNs), which have a computational complexity linear to the number of edges $|E|$.

The encoder generates the representation for each node. When initializing the node embeddings, we use an MLP to obtain a time embedding for the current step $t$ and combine it with the node attributes. By introducing the time step information $t$ which conditions the encoder at different noise levels, we can improve the denoising performance at different diffusion steps. The node representation of \yao{each node $j$} is updated through several layers of message passing: 
\[
H_j^{l+1} = \sigma\left( W^{l}_h H_j^{l} + \sum_{i \in \mathcal{P}j} \frac{1}{|\mathcal{P}(j)|} W^{l}_m H_i^{l} \right)
\]
where $H_j^{l}$ is the representation of node $j$ at encoder layer $l$, $\mathcal{P}(j)$ denotes the parents of node $j$, $W^{l}_h$ and $W^{l}_m$ are learnable weight matrices \yao{of the MPNN-based encoder}, and $\sigma$ is Relu function.

\subsection{{Decoder Design}}
In the denoising model $\phi_{\theta, t}$, the decoder \yao{will decide} the existence of directed edges between pairs of nodes based on their encoded representations $H_i$ from the encoder. In the context of directed graphs, conventional symmetric operations are insufficient to distinguish the directionality of edges. For example, the dot product~\cite{li2023graphmaker} or \yao{Euclidean distance}~\cite{zhang2019d} \yao{of node embeddings} $H_i$, $H_j$ are commutative (i.e., $H_i^{\top} H_j = H_j^{\top} H_i$ and $\left \| H_i -H_j \right \| _2$ = $\left \| H_j -H_i \right \| _2$), failing to differentiate an \yao{directed edge from node $i$ to $j$ or from node $j$ to $i$}.

To address this limitation, we model the directed edge $e_{i,j}$ by associating $H_i$ and $H_j$ with a \yao{learnable} relation embedding $r(t)$~\cite{bordes2013translating}, capturing the inherent directionality of the edge. In an asymmetric relation, the fundamental idea is that the embedding of the source node $H_i$, when translated by the relation embedding $r(t)$,  is supposed to be close to the embedding of the target node $H_j$. We define the decoder's prediction for the existence probability of $e_{i,j}$ at time step $t$ as follows:
\begin{equation}
\begin{split}
P_E^{(t-1)}(i,j)=\hat{p}_{ij}^{(t-1)} &= p_{\theta}\left( A_{ij}^{(t-1)} = 1 \mid H_i^{(t)}, H_j^{(t)}, t \right)\\ &= \mathrm{MLP}\left(\left \{ \left( H_i^{(t)} + r(t) \right) \odot H_j^{(t)}\right \}  \oplus d(t) \right) \notag
\end{split}
\end{equation}
where $H_i^{(t)}$ and $H_j^{(t)}$ are the representations of nodes $i$ and $j$ at time step $t$ from the encoder, $r(t)$ is the learnable relation embedding obtained by $r(t)=MLP_r(t)$, $\odot$ denotes element-wise multiplication, $\oplus$ denotes vector concatenation, $d(t)$ is the learnable embedding of the time step $t$ obtained by $d(t)=MLP_d(t)$.

\begin{figure*}[t]  
    \centering        
    \includegraphics[width=0.95\textwidth]{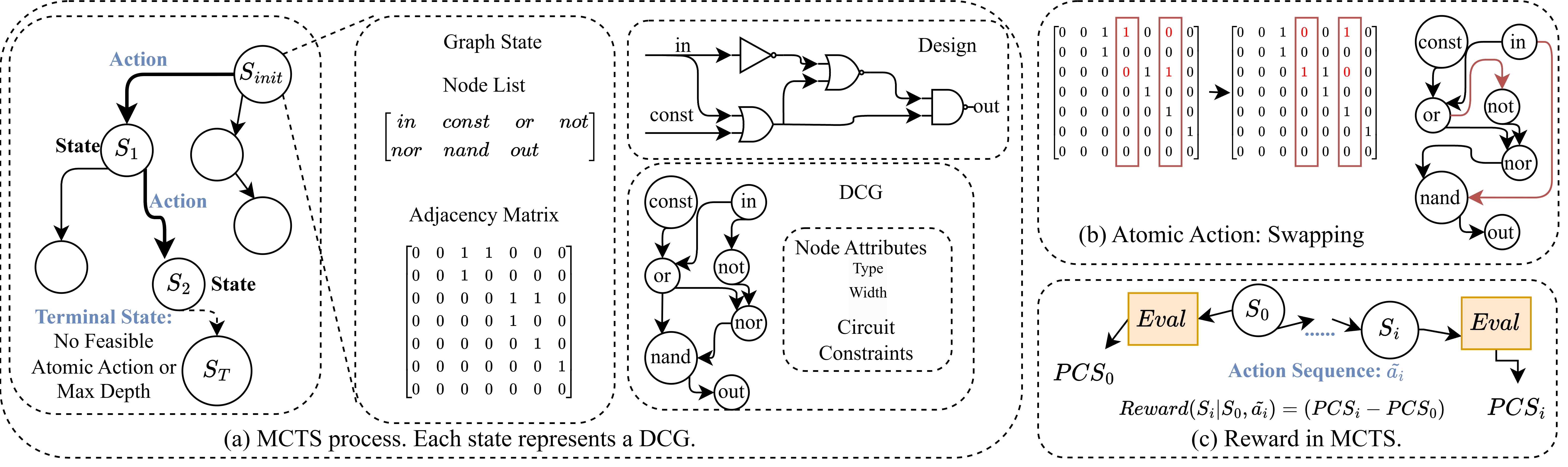}  
    \caption{MCTS-based circuit optimization flow overview. (a) The search tree is on the left and each state node represents an adjacency matrix. (b) A swapping atomic operation on the adjacency matrix facilitates state transformations during the search. (c) We propose a new metric, post-synthesis size (PCS) as the reward model to guide the search process.} 
    \vspace{-.1in}
    \label{fig:MCTS-flow}  
\end{figure*}
\section{Phase 2. Probability-Guided Graph Post-processing}


In phase 1, we obtain the $G^{ini}$ and the edge existence probability $P_E^{(t=0)}$. But the $G^{ini}$ will mostly likely violate circuit constraints $\mathcal{C}$. Therefore, we need to utilize the edge existence probability $P_E^{(t=0)}$ to refine $G^{ini}$ and produce designs $G^{val}$ that meet all predefined circuit constraints.  We propose a probability-guided sequential post-processing approach to leverage the $P_E^{(t=0)}$ for parent node selection while ensuring both fanin limitations and combinational loop prevention which are required by constraints $\mathcal{C}$. 

The post-processing algorithm iteratively processes each node in the circuit graph. For each node $i$, if its parent edges in $G^{ini}$ are valid, then we will skip this node. If this node violates $\mathcal{C}$, we leverage the predicted edge probabilities from the previous diffusion model, sorting potential parent nodes by their connection probabilities in descending order. Before adding an edge from a candidate parent $j$, we need to check if the combinational loops would be created by this new edge. It can be achieved by checking if there exists a path from $i$ to $j$ in the subgraph that excludes register-type nodes. This sequential processing continues until the required number of valid parents is found. 




\section{Phase 3. Refinement on Circuit Redundancy}
\label{refine:red}

\textbf{Logic redundancy problem.}  Another important yet rarely studied challenge is logic redundancy.
In real circuits, human designers will not likely introduce too much redundant logic into the circuit, which can be reflected by the sequential cell count in the optimized netlist. We define a metric, sequential cell preservation ratio (SCPR) which is calculated by dividing the number of sequential cells in the synthesized netlist by the total number of bits in sequential signals in the pre-synthesis HDL design. Our experiments show that the SCPR is usually between 70\% to 100\% in real designs. But some synthetic designs $G^{val}$ can not even reach a 10\% SCPR, which shows a serious logic redundancy problem.

\textbf{Our objective in \circled{3}.} To better align the generated circuits with real ones not only in the topological aspects but also in the redundancy perspective, we aim to refine $G^{val}$ and obtain $G^{opt}$, thus reducing the ratio of removed registers by the synthesis tool. 

\subsection{Challengs and optimization overview.}
 To optimize synthesized circuit designs, we face the challenges of an immense exploration space. To tackle this issue, we optimize the driving cone for the target register\footnote{ The term ``driving cone for a register" refers to the set of nodes obtained by performing a reverse breadth-first search starting from a register node. This search traces back through the parent nodes until nodes of type ``const", ``in", or other ``reg" nodes are encountered. The nodes traversed during this process collectively constitute the driving cone for our target particular register.} one by one. In the cone optimization procedure, we introduce an MCTS-based effective strategy for navigating the large design space and finding circuits with better logic preservation quality. 

\subsection{Monte Carlo tree search-based exploration.} We detail the MCTS process in Figure~\ref{fig:MCTS-flow}. It guides the exploration and optimization of the design space, driving the search toward solutions with reduced hardware redundancy.  Within this framework, each state node represents an adjacency matrix. We define a swapping atomic operation on the adjacency matrix, facilitating state transformations during the search.



\textbf{Hardware redundancy metric as MCTS reward}. To reflect the cone redundancy, we defined the post-synthesis circuit size (PCS). It is calculated by dividing the post-synthesis area by the number of nodes before synthesis. The rationale is that a larger PCS indicates fewer components have been optimized out, implying less redundancy in the synthetic circuit design. The overarching goal of the MCTS is to maximize the PCS while preserving the validity of the circuit.  




\textbf{Action space.} The action space for transforming the circuit design during the MCTS process is defined by our proposed atomic swapping operation. For State $S_i$ with $A(i,j)=1$ and $A(p,q)=1$, the new state $S_{i+1}$ can be obtained by simply swapping the parents of node $j$ and $q$, i.e., $S_{i+1}$ with $A(p,j)=1$ and $A(i,q)=1$. Each atomic action should be checked if it violates $\mathcal{C}$. The advantage of this atomic operation lies in its ability to ensure that the number of edges in the graph remains unchanged, while also maintaining the out-degree and in-degree of each node in $G^{val}$. 

\textbf{Selection and expansion.} In the selection phase, the MCTS algorithm traverses down the tree, selecting the most promising state which also meets circuit constraints $\mathcal{C}$ to expand based on the UCB1 criterion~\cite{browne2012survey}:
\[
a = \arg\max_a \left[ Q(S, a) + \sqrt{2}  \cdot \sqrt{\frac{\ln N(S)}{N(S, a)}} \right]
\]
where \( Q(S, a) \) is the empirical mean reward of action \( a \) in state \( S \), \( N(S) \) is the total number of visits to state \( S \), \( N(S, a) \) is the number of times action \( a \) has been chosen in state \( S \). This formula balances the exploitation of known high-search reward actions with the exploration of less-tried actions. During the expansion step, the MCTS algorithm generates new child nodes by applying the atomic operation to the current node. 

\textbf{Simulation and backpropagation.} Due to our objective of identifying the optimal state encountered during the search trajectory rather than optimizing for terminal states, we modify the traditional MCTS simulation and backpropagation mechanisms. In our approach, the reward for each simulation is defined as the maximum state reward $Reward_{max}$ encountered during that search path, rather than the terminal state value. During backpropagation, we update $Q(S, a)$~\cite{browne2012survey} using the $Reward_{max}$.

\section{Experimental Results} \label{sec:expr}
\textbf{Application of SynCircuit in our experiments.} The register-transfer level (RTL) stage offers maximum optimization flexibility in VLSI flow. Recently, machine learning-based RTL-level Power, Performance, and Area (PPA) prediction methods have been proposed, which can directly predict the performance of designs without requiring logic synthesis~\cite{xu2022sns}~\cite{fang2023masterrtl}~\cite{fang2024annotating}, thereby significantly accelerating the design iteration process. However, the application of these methods is limited by the insufficient availability of open-source RTL code.

\textbf{Experiments objectives.} In this experiment, our primary objective is to explore the effectiveness of new synthetic circuits for PPA modeling at the RTL level. Additionally, we are also interested in investigating the differences between synthetic circuits and real designs from not only the graph structure but also the logic redundancy perspectives. The comprehensive evaluation for SynCircuit may help us gain more insights into the digital circuit generative model.

\subsection{Experimental Setup}

\textbf{Circuit design preparation.} Firstly, we developed a dataset comprising 22 designs based on open-source RTL. The detailed dataset information is provided in Table~\ref{tbl:dataset}. It encompasses a diverse range of digital circuit modules, representing some of the high quality designs available within the open-source community.

\begin{table}[h]
\renewcommand{\arraystretch}{1.2}
\resizebox{0.46\textwidth}{!}{\begin{tabular}{c|c|c|c}
\hline
\textbf{\begin{tabular}[c]{@{}c@{}}Source \\ Benchmark\end{tabular}} & \textbf{\begin{tabular}[c]{@{}c@{}}\#. \\ of Designs\end{tabular}} & \textbf{\begin{tabular}[c]{@{}c@{}}Original \\ HDL Type\end{tabular}} & \textbf{\begin{tabular}[c]{@{}c@{}}Design Scale (\#K Gates) \\ \{Min, Median, Max\}\end{tabular}} \\ \hline
ITC'99\cite{corno2000rt}                                  & 6                                                                  & VHDL                                                                  & \{9, 19, 45\}                                                                                     \\
OpenCores\cite{albrecht2005iwls}                               & 8                                                                  & Verilog                                                               & \{2, 6, 35\}                                                                                      \\
Chipyard\cite{amid2020chipyard}                                & 8                                                                  & Chisel                                                                & \{12, 19, 52\}                                                                                    \\ \hline
\end{tabular}
}
\caption{Dataset composition and design size information.}
\label{tbl:dataset}
\end{table}

\begin{table*}[t]
\renewcommand{\arraystretch}{1.2}
\centering
\resizebox{0.95\textwidth}{!}{%
\begin{tabular}{lclclcllclclcl}
\hline
\multirow{3}{*}{Model} & \multicolumn{6}{c}{1-Wasserstein distance $W_1 \downarrow$}                                                                                           &  & \multicolumn{6}{c}{$\mathbb{E}_{\hat{G} \sim p_{\theta}} \left[ \frac{M(\hat{G})}{M(G)} \right] \rightarrow 1$}       \\ \cline{2-7} \cline{9-14} 
                       & \multicolumn{2}{c}{Out Degree} & \multicolumn{2}{c}{Cluster} & \multicolumn{2}{c}{Orbit}             &  & \multicolumn{2}{c}{\# Triangle}       & \multicolumn{2}{c}{$\hat{h}(A, Y)$}   & \multicolumn{2}{c}{$\hat{h}(A^2, Y)$} \\
                       & \multicolumn{1}{l}{TinyRocket}    & Core   & \multicolumn{1}{l}{TinyRocket}   & Core   & \multicolumn{1}{l}{TinyRocket} & Core &  & \multicolumn{1}{l}{TinyRocket} & Core & \multicolumn{1}{l}{TinyRocket} & Core & \multicolumn{1}{l}{TinyRocket} & Core \\ \cline{1-7} \cline{9-14} 
GraphRNN~\cite{you2018graphrnn}                     & 1.03                        & 0.762       &0.935                  &0.136        &1.94                   &0.978      &  &0.016               &0.0238      &5.33            &3.27      &2.13            &2.53      \\
DVAE~\cite{zhang2019d}                    & 1.31                            &0.832        & 0.912                   &0.146      & 1.33                  &1.24       &  & 0.241            &0.105      & 6.66                      &5.19      & 3.17                      &2.96      \\
GraphMaker-v~\cite{li2023graphmaker}                    &0.678                             &0.621        &0.957                    &0.112        &1.24                  &1.05      &  &\textbf{0.262}             &0.501      & 2.31               &1.96      &1.53                      &1.62      \\
Sparse Digress-v~\cite{qin2023sparse}                   &0.598                      &0.652        &0.972                    & 0.135       &1.21                  &0.953      &  &0.163             &0.312      &4.89               &4.32      &2.56                       &2.13      \\ \hline
SynCircuit w/o diff                   &0.373                                & 0.323       &0.925                               &0.0807        &1.09                             &0.926      &  &0.0501                             &\textbf{0.760}      &0.629                             &0.361      &0.561                             &0.321      \\ 
SynCircuit  w/ diff           & \textbf{0.236}                     &\textbf{0.226}        &\textbf{0.876}                  &\textbf{0.0452}        & \textbf{0.344}                  &\textbf{0.231}   &   & 0.146              &1.34      &\textbf{0.713}                      &\textbf{0.670}      &\textbf{0.624}                      & \textbf{0.487}      \\ \hline
\end{tabular}
}
\caption{Evaluation for structural properties similarity with original realistic circuits, best results are in \textbf{bold}. Lower $W_1$ and $\left | \mathbb{E}_{\hat{G} \sim p_\theta} \left[ M(\hat{G})/{M(G)} \right]-1 \right |$  reflect a better graph similarity with the realistic ones. All 4 baselines were adapted to generate directed circuit graphs without violating the constraints.}
\vspace{-.1in}
\label{tbl:strcture}
\end{table*}

\textbf{Design label preparation.} To obtain netlist labels including design area, register slack (SL), worst negative slack (WNS), and total negative slack (TNS), we employed Synopsys Design Compiler® 2021 with the NanGate 45nm technology library. To align with real-world scenarios, multiple parameters within the Design Compiler were adjusted, and a set of the  PPA values along the Pareto frontier were utilized as ground truth labels.

\textbf{Training-testing data splitting.} We randomly selected 7 designs from the dataset as the test set, while the remaining 15 designs were used as the training set in the downstream PPA prediction task. To prevent data leakage, all the graph generative models mentioned in this paper were only trained on these 15 training designs.

\textbf{Graph generative model baselines.} We selected several representative and most closely related baselines, including GraphRNN~\cite{you2018graphrnn}, DVAE~\cite{zhang2019d}, SparseDigress~\cite{qin2023sparse} and GraphMaker~\cite{li2023graphmaker}. These 4 baselines all need to be adapted to the circuit generation task.  GraphRNN~\cite{you2018graphrnn} and DVAE~\cite{zhang2019d} are node ordering-based autoregressive generative approaches. These methods cannot be directly applied to generate directed cyclic graphs. We have to break the cycles in the training circuits and use the topological order of nodes as the sequence for the model training and inference. During generation, edge directions are automatically determined by the topological order.  In the sequential generation process, we introduced a validity checker for circuits to ensure that the resulting digital design is valid. \textbf{But since these two models can only be applied on DAGs, the generated graph contains no cycles which is very different from the real designs.}



For GraphMaker~\cite{li2023graphmaker} and SparseDigress~\cite{qin2023sparse}, these are one-shot approaches that ignore edge direction information, generating only undirected graphs. We applied the Gravity-Inspired Graph Autoencoders~\cite{salha2019gravity} method to process the generated undirected graph by assigning the direction for each edge. 
To ensure the validity of the generated graphs by GraphMaker~\cite{li2023graphmaker} and SparseDigress~\cite{qin2023sparse}, we must refine the parent edges in a specific node order. The synthesis tool removed most part of the generated circuits, so these two baseline models can only be used for comparing graph characteristics and cannot support valid designs for downstream tasks.

\textbf{SynCircuit setup.} For our proposed graph generative model, we set the diffusion steps to 9 and utilized a 5-layer MPNN. Both the node attribute embeddings and hidden embeddings were set to 256. In the circuit optimization phase, to accelerate the evaluation process, we replaced the slow synthesis tool with a trained discriminator to approximate the PCS. In the MCTS process, we set the number of simulations to 500 and the maximum exploration depth to 10 for each register cone. All these experiments were conducted on a platform with an Intel(R) Xeon(R) Gold 6438Y+ processor and 8*4090 GPUs.

\begin{table*}[]
\renewcommand{\arraystretch}{1.3}
\centering
\resizebox{0.98\textwidth}{!}{
\begin{tabular}{llccclccccccclccc}
\hline
\multirow{2}{*}{Model} &  & \multicolumn{3}{c}{Register Slack}            &  & \multicolumn{3}{c}{WNS}                       &  & \multicolumn{3}{c}{TNS}                       & \multicolumn{1}{c}{} & \multicolumn{3}{c}{Area}                      \\ \cline{3-5} \cline{7-9} \cline{11-13} \cline{15-17} 
                       &  & R $\rightarrow 1$             & MAPE $\downarrow$          & RRSE $\downarrow$          &  & R $\rightarrow 1$             & MAPE $\downarrow$          & RRSE $\downarrow$          &  & R $\rightarrow 1$             & MAPE $\downarrow$          & RRSE $\downarrow$          &                      & R $\rightarrow 1$             & MAPE $\downarrow$          & RRSE $\downarrow$          \\ \hline
Basic training data (No-pseudo circuits)     &  & 0.70           & 27\%          & 0.83          &  & 0.86          & 20\%          & 0.83          &  & 0.81          & 50\%          & 0.97          &                      & 0.89          & 30\%          & 0.62          \\ \hline
GraphRNN~\cite{you2018graphrnn}               &  & 0.70           & 27\%          & 0.83          &  & 0.88          & 21\%          & 0.83          &  & 0.80          & 54\%          & 0.97          &                      & 0.84          & 44\%          & 0.75          \\
DVAE~\cite{zhang2019d}                   &  & 0.69          & 29\%          & 0.94          &  & 0.88          & 24\%          & 0.86          &  & 0.78          & 50\%          & 0.97          &                      & 0.84          & 61\%          & 0.86          \\ \hline
SynCircuit w/o opt     &  & 0.72          & 24\%          & 0.79          &  & \textbf{0.90} & 23\%          & 0.85          &  & 0.80          & 48\%          & 0.86          &                      & 0.86          & 39\%          & 0.72          \\
SynCircuit w/ opt      &  & \textbf{0.77} & \textbf{16\%} & \textbf{0.70} &  & 0.89          & \textbf{20\%} & \textbf{0.80} &  & \textbf{0.97} & \textbf{45\%} & \textbf{0.64} &                      & \textbf{0.95} & \textbf{25\%} & \textbf{0.34} \\ \hline
\end{tabular}
}
\caption*{(a) Basic training dataset contains 15 real designs. Lower $\left | R-1 \right |$, MAPE and RRSE reflect better model prediction performance.}
\end{table*}

\begin{table*}[!t]
\renewcommand{\arraystretch}{1.3}
\centering
\resizebox{0.98\textwidth}{!}{
\begin{tabular}{llccclccccccclccc}
\hline
\multirow{2}{*}{Model}                                &  & \multicolumn{3}{c}{Register Slack}            &  & \multicolumn{3}{c}{WNS}                      &  & \multicolumn{3}{c}{TNS}                       & \multicolumn{1}{c}{} & \multicolumn{3}{c}{Area}                      \\ \cline{3-5} \cline{7-9} \cline{11-13} \cline{15-17} 
                                                      &  & R $\rightarrow 1$             & MAPE $\downarrow$          & RRSE $\downarrow$          &  & R $\rightarrow 1$             & MAPE $\downarrow$          & RRSE $\downarrow$         &  & R $\rightarrow 1$             & MAPE $\downarrow$          & RRSE $\downarrow$          &                      & R $\rightarrow 1$             & MAPE $\downarrow$          & RRSE $\downarrow$          \\ \hline
Basic training data (No-pseudo circuits)                                    &  & 0.52          & 34\%          & 1.1           &  & NA            & 52\%          & 2.1          &  & NA            & 67\%          & 1.1           &                      & 0.65          & 66\%          & 1.3           \\ \hline
GraphRNN\cite{you2018graphrnn} &  & 0.52          & 34\%          & 1.1           &  & 0.71          & 42\%          & 1.7          &  & -0.30          & 74\%          & 1.1           &                      & 0.51          & 77\%          & 1.6           \\
DVAE\cite{zhang2019d}          &  & 0.49          & 36\%          & 1.3           &  & 0.75          & 77\%          & 2.6          &  & 0.76          & 93\%          & 1.1           &                      & 0.70          & 86\%          & 2.4           \\ \hline
SynCircuit w/o opt                                    &  & 0.56          & 35\%          & 1.2           &  & 0.65          & 47\%          & 1.9          &  & 0.72          & 70\%          & 0.85          &                      & 0.63          & 62\%          & 1.1           \\
SynCircuit w/ opt                                     &  & \textbf{0.67} & \textbf{25\%} & \textbf{0.76} &  & \textbf{0.87} & \textbf{34\%} & \textbf{1.2} &  & \textbf{0.98} & \textbf{61\%} & \textbf{0.63} &                      & \textbf{0.95} & \textbf{36\%} & \textbf{0.61} \\ \hline
\end{tabular}

}
\caption*{(b) Basic training dataset contains 5 real designs. Lower $\left | R-1 \right |$, MAPE and RRSE reflect better model prediction performance.}
\caption{Model Performance on the Register Slack, WNS, TNS, and area prediction.  The basic training dataset in (a) and (b) contains 15 and 5 real designs respectively. In (a) and (b), the augmentation datasets are added to the basic training set, each always with 25 pseudo-circuits, generated from SynCircuit, GraphRNN~\cite{you2018graphrnn}, and DVAE~\cite{zhang2019d}. R with NA means that the model prediction is the same for all testing designs}
\vspace{-.15in}
\label{tbl:model-performance}
\end{table*}
\vspace{-.1in}

\subsection{Experimental Results}

\subsubsection{Graph structural properties evaluation}
To evaluate the similarity of the generated graph structures with realistic ones, we follow GraphRNN~\cite{you2018graphrnn} and GraphMaker~\cite{li2023graphmaker} and report distance metrics for node out degree, clustering coefficient, and four-node orbit count distributions. We use the 1-Wasserstein distance to model statistic distributions from the original and generated graphs, respectively.  A lower $W_1$ value indicates better similarity.

Beyond distribution distance metrics, we also directly compare several scalar-valued statistics~\cite{li2023graphmaker, lim2021large}. Let $M(G)$ be a non-negative statistic; we report $\mathbb{E}_{\hat{G} \sim p_\theta} \left[ M(\hat{G})/{M(G)} \right]$, where $\hat{G}$ is a generated graph. A value closer to 1 indicates better performance. For the $M$ statistics, we choose triangle count, $\hat{h}(A, X)$~\cite{li2023graphmaker, lim2021large} and two-hop correlations $\hat{h}(A^2, X)$. $\hat{h}(A, X)$ reflects the correlation between graph structure and node types.

The results are shown in Table~\ref{tbl:strcture}. We can see that our SynCircuit has the best performance among 5 out of the 6 metrics. Benefiting from the diffusion model and the out-degree guidance in the postprocessing phase, the degree distribution of our synthetic circuits is more similar to the real designs. This feature is important because the digital circuits are indeed scale-free networks\footnote{A Scale-Free Network is a type of network characterized by a degree distribution that follows a power law. This means that in such networks, the probability $P(k)$ that a node has $k$ connections (or degree $k$) decreases polynomially as $k$ increases}~\cite{wang2003complex}.   

SynCircuit w/o diff is an ablation study where we remove the diffusion model and randomly construct edges when generating $G^{ini}$ and $P_E^{(0)}$, but applying post-processing to ensure certain predefined constraints. We observe that its performance is greatly inferior to the complete SynCircuit, demonstrating the effectiveness of our designed generative diffusion model.

\subsubsection{Logic redundancy and timing evaluation}

\begin{figure}[h]  
    \centering        
    \includegraphics[width=0.5\textwidth]{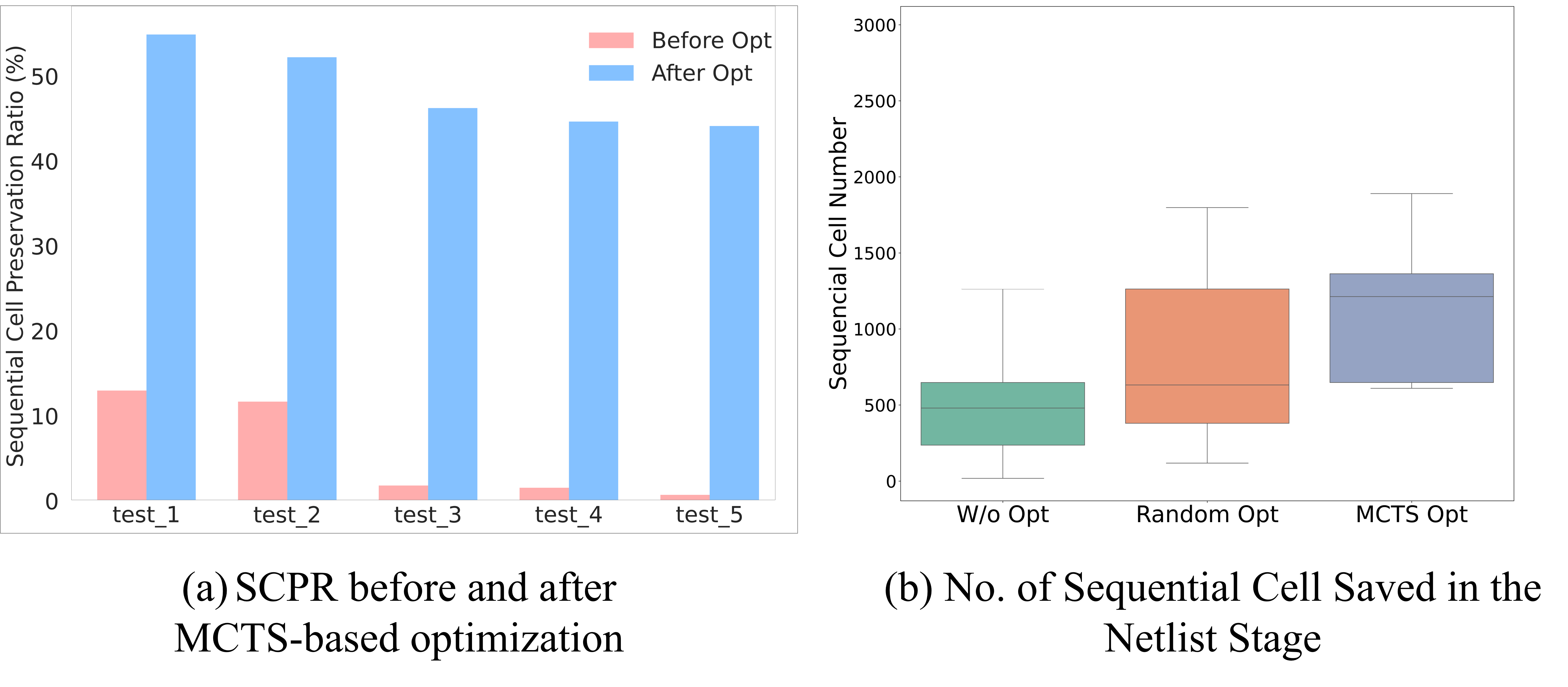}  
    \caption{(a) Logic redundancy metric SCPR is calculated by dividing the number of sequential cells in the synthesized netlist by the total number of bits in sequential signals in the pre-synthesis HDL design. Here we select the five $G^{val}$ examples with the most significant logic redundancy. The register ratio (SCPR) cannot even reach 20\% if we do not optimize $G^{val}$. After the MCTS optimization, the SCPR is greatly improved and exceeds 50\% in some synthetic circuits. (b) The number distribution of registers preserved after logic synthesis. The logic redundancy is greatly reduced by our proposed MCTS optimization method.}  
    \label{fig:mcts-performance}  
\end{figure}

We employ Monte Carlo Tree Search (MCTS) to optimize the register cones within $G^{val}$. For comparison, we implement an ablation study of randomly altering edge connections on $G^{val}$ while still ensuring every step is valid. We utilize the same number of simulations as MCTS and adopt the optimal solution identified throughout the process as the results. The SCPR enhancement is illustrated in Figure~\ref{fig:mcts-performance}(a), demonstrating that MCTS can significantly reduce logic redundancy.

Figure~\ref{fig:mcts-performance}(b) displays the number of sequential cells saved during the logic synthesis using different optimization techniques. We can see that by introducing our MCTS-based refinement, the registers saved in the netlist have been greatly improved over no optimization and the optimization performance is also better than that using a random optimization method.

\textbf{Netlist timing statistics comparison}
We use two metrics to evaluate the design's timing characteristics: Worst Negative Slack (WNS) and Total Negative Slack divided by the number of violated paths (TNS/NVP). 
\begin{itemize}
    \item WNS represents the most severe timing violation in the design, indicating the largest negative slack among all timing paths, which points to the longest delay and the most critical bottleneck in signal propagation.
    \item The metric TNS/NVP provides the average negative slack per violated path, reflecting the overall severity of timing violations and offering insights into the distribution of delays.
\end{itemize}

\begin{figure}[h]  
    \centering        
    \includegraphics[width=0.5\textwidth]{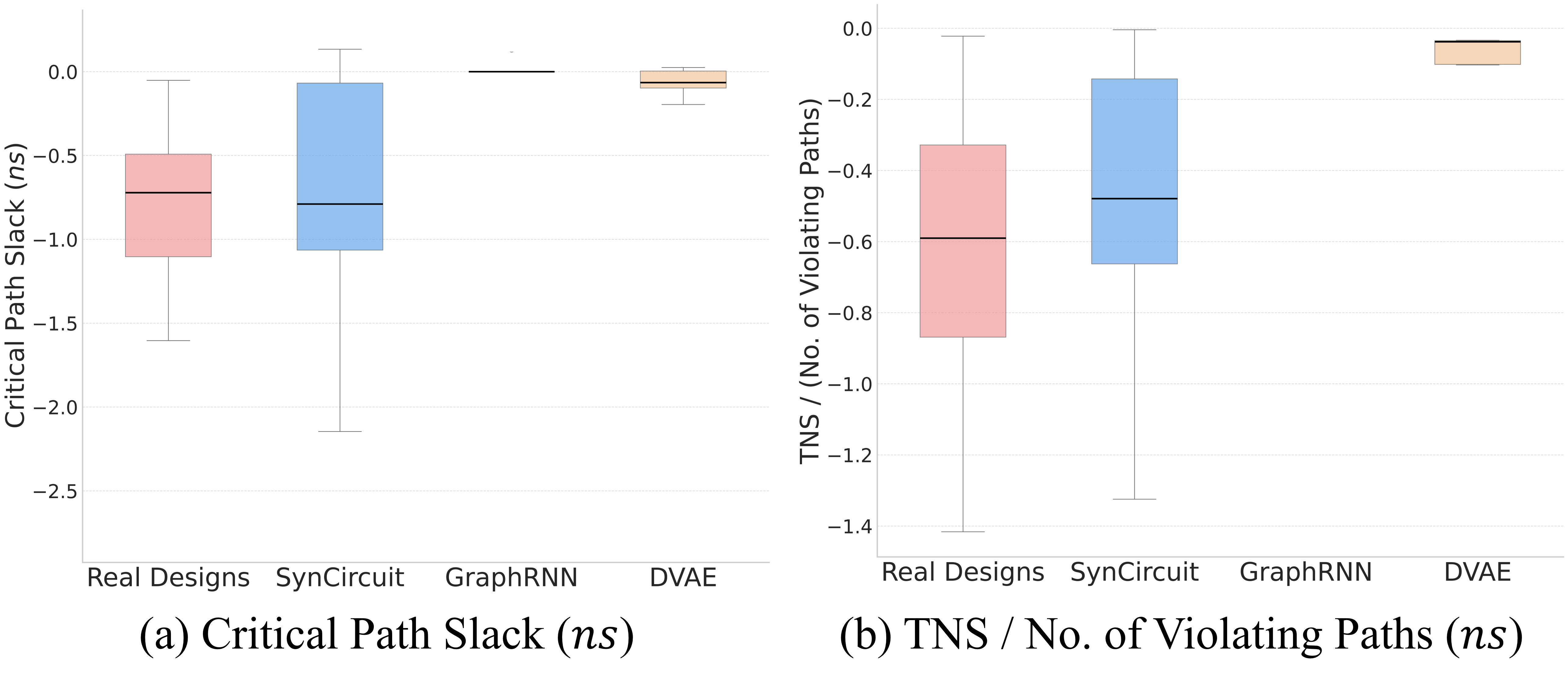}  
    \caption{Netlist statistics for the three synthetic datasets and real benchmarks. The distributions of two statistic metrics: Critical Path Slack (WNS) and the ratio of Total Negative Slack to the Violating Path Numberare shown in (a) and (b), respectively. The dataset generated by SynCircuit exhibits \emph{more similar statistics distributions} to those of the real designs compared to GraphRNN~\cite{you2018graphrnn} and DVAE\cite{zhang2019d}.}  
    \label{fig:netlist-box} 
    \vspace{-.1in}
\end{figure}

From Figure~\ref{fig:netlist-box}, We observe that the graphs generated by GraphRNN~\cite{you2018graphrnn} and DVAE~\cite{zhang2019d} exhibit very small WNS (critical path slack) and TNS/NVP values, failing to capture the inherent delay characteristics of circuits. In contrast, our SynCircuit demonstrates a more similar distribution of the two metrics to real designs. This suggests that SynCircuit is more effective in modeling the diverse timing behaviors present in real circuits.

\subsubsection{SynCircuit application in RTL tasks}
In this part, we explore the application of synthetic circuit generation for ML-based RTL-level PPA prediction model, primarily referencing the overall design evaluation methods (i.e., area, WNS, and TNS) proposed by MasterRTL~\cite{fang2023masterrtl} and the fine-grained timing slack evaluation by RTL-Timer~\cite{fang2024annotating}. We employ three metrics to assess model performance: correlation coefficient ($R$), Mean Absolute Percentage Error (MAPE), and Root Relative Squared Error (RRSE). Lower MAPE and RRSE values indicate better model performance.

We randomly selected 5 and the full 15 training real designs to create two basic training datasets. For each basic training dataset, we augmented it with different synthetic datasets (three sets of 25 designs each), generated respectively by SynCircuit, GraphRNN~\cite{you2018graphrnn}, and DVAE~\cite{zhang2019d}, to study how these synthetic designs affect model performance.

The results are shown in Table~\ref{tbl:model-performance}. We can see that in both training settings, models trained on datasets augmented with Synthetic-generated data consistently outperformed those trained solely on real designs, achieving the best results across all metrics. And we have more model performance gain when the basic training dataset contains 5 designs. The Register Slack MAPE is reduced by 10\% in both basic training settings. And we have a 30\% area MAPE reduction in the 5 basic training dataset settings.

Notably, models augmented with DVAE~\cite{zhang2019d}-generated and GraphRNN~\cite{you2018graphrnn}-generated data performed worse, regardless of the training dataset size. This may indicate a significant gap between the data generated by these baselines and the real designs due to logic redundancy. As observed in Figure~\ref{fig:netlist-box}, the synthetic data generated from GraphRNN~\cite{you2018graphrnn} and DVAE~\cite{zhang2019d} contain very few paths with large delays. This discrepancy may have caused the model's learning to deviate from the normal timing features for Register Slack, WNS, and TNS tasks.

We also included non-optimized circuits SynCircuit w/o opt ($G^{val}$) as part of an ablation study. It can be observed that the excessive design logic redundancy introduced may have adversely affected the model's predictive performance, rendering it inferior to scenarios involving only real training circuits. This observation indirectly demonstrates the importance of optimizing logic redundancy in generating synthetic circuits that more closely resemble real-world designs. 
\section{Conclusion And Future Work}
Data-driven automation techniques have become increasingly prevalent in digital circuit design in recent years. Nevertheless, the availability of open-source circuits is often limited. To address this issue, we propose the SynCircuit, an automatic digital circuits generation framework. It consists of three stages: directed cyclic graph generation, probability-guided postprocessing, and MCTS-based optimization.  Our experiments demonstrate that the designs generated by SynCircuit not only exhibit structural properties that closely resemble those of real designs, but also enhance model performance in ML-based power, performance, and area (PPA) prediction tasks at the RTL early stage. 
\section{Acknowlegdement}
This work is supported by the Hong Kong Research Grants Council (RGC) ECS Grant 26208723, CRF Grant C6003-24Y, and ACCESS – AI Chip Center for Emerging Smart Systems, sponsored by InnoHK, Hong Kong SAR. We thank HKUST Fok Ying Tung Research Institute and the National Supercomputing Center in Guangzhou Nansha Sub-center for computational resources.
\bibliographystyle{IEEEtran}
\bibliography{ref}

\end{document}